**Research Article**

**A fall alert system with prior-fall activity identification**

Pisol Ruenin[1], Sarayut Techakaew[1], Patsakorn Towatrakool[2] and Jakarin Chawachat[1,*]

[1]Department of Computer Science, Faculty of Science, Chiang Mai University, Chiang Mai, 50200, Thailand.

[2]Department of Computer Engineering, Faculty of Engineering, Chiang Mai University, Chiang Mai, 50200, Thailand.

[*]Corresponding author. Email address: jakarin.c@cmu.ac.th

**Abstract**

Falling, especially in the elderly, is a critical issue to care for and surveil. There have been many studies focusing on fall detection. However, from our survey, there is still no research indicating the prior-fall activities, which we believe that they have a strong correlation with the intensity of the fall. The purpose of this research is to develop a fall alert system that also identifies prior-fall activities. First, we want to find a suitable location to attach a sensor to the body. We created multiple-spot on-body devices to collect various activity data. We used that dataset to train 5 different classification models. We selected the XGBoost classification model for detecting a prior-fall activity and the chest location for use in fall detection from a comparison of the detection accuracy. We then tested 3 existing fall detection threshold algorithms to detect fall and fall to their knees first, and selected the 3-phase threshold algorithm of Chaitep and Chawachat [3] in our system. From the experiment, we found that the fall detection accuracy is 88.91%, the fall to their knees first detection accuracy is 91.25%, and the average accuracy of detection of prior-fall activities is 86.25%. Although we use an activity dataset of young to middle-aged adults (18-49 years), we are confident that this system can be developed to monitor activities before the fall, especially in the elderly, so that caretakers can better manage the situation.

## 1. Introduction

Nowadays, the world's population is ageing. Care and surveillance for the elderly are one of the critical issues in preparing to become an ageing society. When looking at the risk of health, falling is one of the leading causes of injury to the elderly. The World Health Organization [15] reported that about 28-35% of the population aged 65 years and older fall each year and rise to 32-42% at the age of 70 years or more. This report is an indication that the frequency of falls increases with age. When an elderly fall, if we can detect it quickly, it will allow a prompt response from the caretaker and minimize other risks that follow.  There exist some devices that allow the wearers to send out calls for help when a fall happens. However, there may be some situations in which an elderly fall and cannot move or lose consciousness, and thus, cannot actively call for help using such method. Hence the need for automatic fall detection is essential.

Another issue to consider when a fall is detected is that the prior-fall activities, the activities that the elderly have been doing right before they fall. These prior-fall activities have a profound effect on the intensity of the injury. For example, if an elderly falls while walking up the stairs, there is a chance that the intensity of injury is more than that of a fall experienced while walking. Therefore, if we can identify the activities involved right before the fall, it will significantly help caretakers manage their responses effectively.

In this research, we developed a fall alert system with prior-fall activity identification. Our system is composed of 2 parts: devices and a server. First, we created devices using a mini Wi-Fi board connected to an accelerometer and gyroscope sensor. We placed the devices on various spots on the subjects' bodies to collect activity data. Then we trained classifiers to detect activities, tested for accuracy, and chose the most accurate detection classifier to use in the server. Moreover, we chose the most accurate spot on the body to test for a fall. We compared three threshold algorithms to find a suitable algorithm and applied the best one as a fall detector in the device. When a fall occurs, the device detects the fall and sends a recorded prior-fall activity to the server. The server processes and classifies prior-activity data and then sends a detected prior-fall activity to caretakers.

The remainder of this paper is organized as follows. Section 2 discusses the previous literature related to activity recognition and fall detection. Section 3 describes our methodology, including data collection, data analysis framework, and a fall alert system with prior-fall activity identification. The experimental results are presented in Section 4. Section 5 and 6 provide discussion and some conclusions, respectively.

## 2. Related Work

Our work consists of two main parts: activity recognition and fall detection. In this section, we discuss these two related research topics.

### 2.1 Activity recognition

Nowadays, wearable sensors are extensively used in activity identification and fall detection. Most of the work was using mobile phones or creating their own devices. In [6, 13], mobile phones were used to store movement data because they have capabilities and sensors in phones like accelerometer sensor, gyroscope sensor, GPS, Light sensor, temperature sensor. After that, mobile phones sent recorded data to the server to detect activities. However, the use of mobile phones has limitations regarding the location of the attachment on the user.

Many researchers have created devices to predict activities. Most devices were equipped with accelerometers [10, 14] or combined with gyroscopes [5]. These sensors capture the subject's movement data. The movement data is then sent to the server or processed by itself. When the server receives the raw signals, it usually processes raw data to get the informative features and then sends it to classification models to predict activities. Recently, Zhang *et al.* [16] used XGBoost to recognize activities. From their experiment, it was found that XGBoost performed better than Random Forest, Gradient Boosted Decision Tree (GBDT), Multi-Layer Perceptron (MLP), SVM, and *k*-NN.

Moreover, most researchers used values from sensors in the Cartesian coordinate system to predict activities. Lehsan and Bootkrajang [7] converted activity data from the Cartesian coordinate system to the Spherical coordinate system. They found that using Spherical coordinate system with activities related to rotating movements yields better detection results.

*2.2 Fall Detection*

Fall detection has long been of interest. Many studies used fall detection thresholds with values measured by accelerometer and gyroscope sensors. Activities are identified as fall if the measured values from sensors meet all the thresholds. Much research used only the values obtained from an accelerometer sensor to detect falls. Some research used a 2-phase threshold to distinguish from ordinary activities, such as the work of Bourke *et al.* [1] and Cao *et al.* [2]. To specify that activity is a fall, at first, the total value acceleration must be lower than the first threshold (Lower Fall Threshold (LFT)), then higher than the second threshold (Upper Fall Threshold (UFT)).

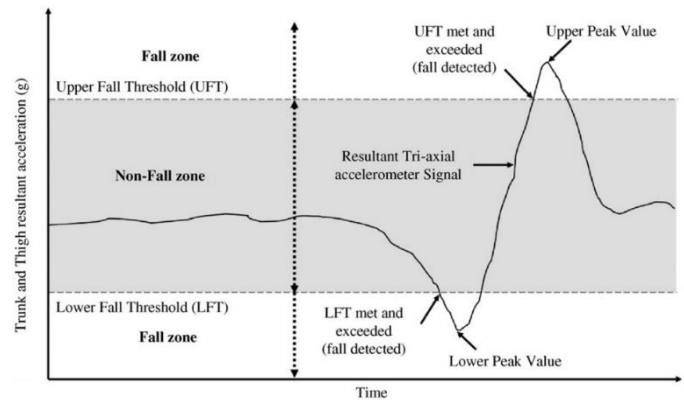

**Figure 1** the 2-phase threshold of [1] using lower and upper fall threshold.

There are also 3-phase threshold algorithms, such as [3, 8]. The 3-phase threshold algorithm is similar to the 2-phase threshold algorithm but adds the final part (phase 3) to guarantee fall, not other activities. The results show that the fall detection is more accurate and less false positive value than the 2-phase threshold algorithms.

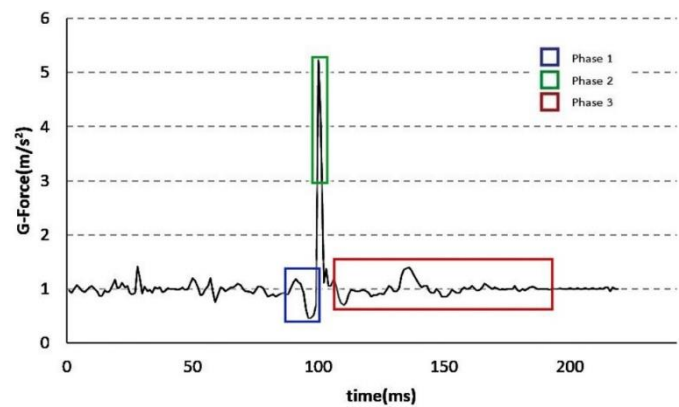

**Figure 2** the 3-phase threshold algorithm proposed by [3].

## 3. Materials and methods

This research consists of 2 main parts, which are data collection and data analysis framework.

### 3.1 Data collection

Initially, we built small devices attached to locations on the subjects' body to collect activity data. We then select the location that predicts the most accurate activity to collect the data of fall and fall to their knee first.

*3.1.1 Devices*

The device should be small, cheap, and widespread for easy re-experimentation and extension. We used WeMos D1 Mini V2, a mini Wi-Fi board based on the ESP-8266EX, connected to the sensor MPU6050, which has accelerometer and gyroscope sensors, and a 3.7V 1800mAh battery as a power supply. This device detects falls and falls to their knees first, then sends the subject movement data via Wi-Fi to the server. Figure 3(a) shows a device connected to a power supply.

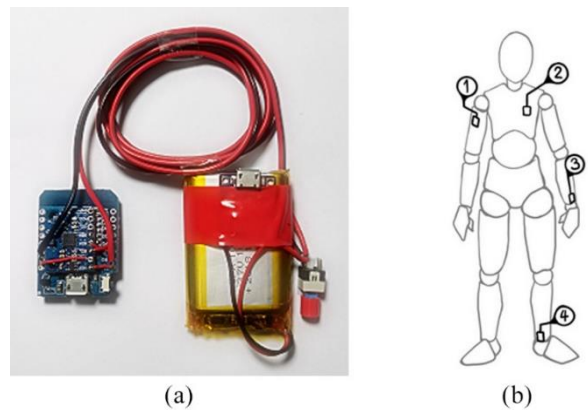

(a)                    (b)

**Figure 3** (a) device connected to a power supply, (b) the 4 sensor locations on the body.

Sensor MPU6050 provides measurements of acceleration ($A_x$, $A_y$, $A_z$) and rotation ($G_x$, $G_y$, $G_z$) around 3-axis. It measures acceleration ranges from $\pm 2g$ to $\pm 16g$ and angular velocity from $\pm 250$ to $\pm 2000º$/sec. From a survey of online activity recognition using mobile phones [12], most research uses 50 Hz sampling rate. Thus, in our research, we used this value as well.

*3.1.2 Activity data collection*

One of the problems with wearable devices for activity recognition is which location on the body can provide the most accurate activity detection. Therefore, we compared the predictive accuracy of body sensor placement in different locations, including right arm (upper arm), left chest (shirt pocket), left wrist, and left foot (ankle), as shown in Figure 3(b). We chose a set of activities that can be done in daily living. The lists of performed and recorded activities are walking, walking upstairs, walking downstairs, jumping, jumping jack, running, sitting, sit up, stand, Sit and then stand up, and Stand and then sit down, which are labelled with WALK, WALK_UP, WALK_DOWN, JUMPING_JACK, JUMP, RUN, SIT, SIT_UP, STAND, UP and DOWN respectively.

We collected activity data from lecturers and students in our laboratory. There were 15 healthy volunteers composed of 13 males and 2 females in the age range 18-49 years and the height range between 150 to 183 cm. All subjects signed a consent form before participating in the experiments. Subjects were asked to perform eleven activities freely and without movement restrictions.

The activity data collection protocol is as follows:

- Install 4 sensors on the subject, as in Figure 3(b).

- Perform each activity for at least 2 minutes, except for JUMPING and JUMPING_JACK, which are performed 30 times instead.

- Repeat at least 3 times for each activity.

*3.1.3 Fall data collection*

After analyzing the activity data, we select the location of the sensor to attach to the body. We collect 2 types of falls: falls and falls to their knee first. Since falls are one of the risky injury activities, we collected fall data from four researchers working on this research. All subjects were males in the age range 18 - 38 years and the height range between 165 to 183 cm. The subjects were collected falls data in 8 directions (North (in front of the subjects), North-East, East, South-East, South (back of the subjects), South-West, West, and North-West) as shown in Figure 4(a) and falls to their knees first in 5 directions (North (in front of the subjects), North-East, East, West, and North-West) as shown in Figure 4(b).

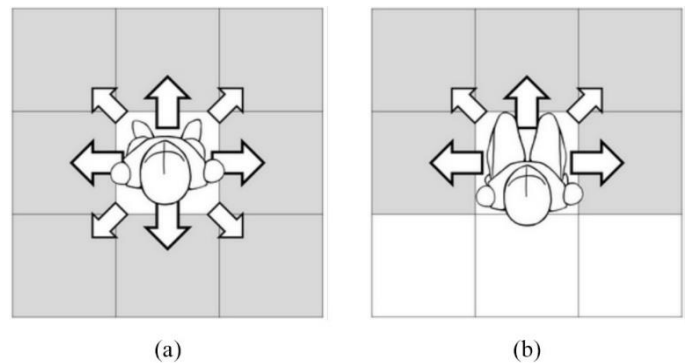

(a)                              (b)

**Figure 4** (a) 8 directions of fall, (b) 5 directions of fall to their knees first.

The protocol for collecting falls and falls to their knees first data is that each subject stand, then falls in each direction 2 times. Both activity and fall dataset can be download at https://www2.cs.science.cmu.ac.th/staff/jakarin/

*3.2 Data analysis framework*

The framework of this work has 3 parts: activity recognition, fall detection, and alert system. In the activity recognition part, we used the collected activity dataset to train classifiers to predict activities. Then, we compared the accuracy of classifiers and chose the most accurate classifier. We also look for the most appropriate sensor location and coordinating system. Since we wanted the device to detect the fall by itself, we used a threshold algorithm that is not complicated to compute. We compared three threshold algorithms. Finally, we combined activity detection and fall detection into a fall alert system with prior-fall activity identification.

*3.2.1 Activity Recognition*

First, we preprocessed the activity dataset from the activity data collection section. We used this dataset to train activity recognition classifiers. Then we compared the accuracy of each classifier.

*3.2.1.1 Preprocess activity data*

Activity data were collected in the form of a data stream. We divided the data stream into 2 -second data windows. Each window consisted of 100 sets of 3-axis accelerometer ($A_x$, $A_y$, $A_z$) and 3-axis gyroscope ($G_x$, $G_y$, $G_z$) in the Cartesian coordinate system. Because we wanted to compare the accuracy of predicting the activities between the Cartesian and Spherical coordinate system in the same way as [7], we also converted an activity data under the Cartesian coordinate system ($x$, $y$, $z$) into another set of data under the Spherical coordinate system ($r$, $\theta$, $\varphi$) by these formulae.

$$r = \sqrt{x^2 + y^2 + z^2},$$

$$\theta = \arccos \frac{z}{\sqrt{x^2 + y^2 + z^2}},$$

$$\varphi = \arctan \frac{y}{x}$$

We selected mean, standard deviation, and Pearson correlation like [7] to be features. We added these features to each data window.

### 3.2.1.2 Compare activity classifiers and sensor locations

We wanted to choose the most accurate classifier to identify activity and then used it in our alert system. We compared the performances from 4 well-known classifiers, namely *k*-Nearest Neighbors, Support Vector Machine, Naive Bayes, and Decision Tree, and XGBoost, which is a popular machine learning. We compared the dataset under the Cartesian and Spherical coordinate system to determine which coordinate system is suitable for predicting activity. We also examined the efficiency of activity detection when sensors are attached in different locations.

### 3.2.2 Fall detection

### 3.2.2.1 Preprocess fall data

Since we do not know in which direction the fall will occur. Therefore, we needed the directionless acceleration for processing. We then combined the values from the 3-axis accelerometer using the Pythagorean Theorem. Finally, we divided that value by 9.8 to get G-Force.

$$Total\ Acceleration = \sqrt{A_x^2 + A_y^2 + A_z^2},$$

$$\text{G-Force} = \frac{Total\ Acceleration}{9.8}$$

We cropped each fall data into 20 dimensions (crop size = 20). First, we selected the highest value as the mark point. To get the right endpoint, we continued to collect data until the value is below average, then continued counting 10 values, or distance from the mark point is equal to 15 values. From the right endpoint, we gathered data back equal to the crop size range to get to the left endpoint. Finally, we filtered out the noise by the Butterworth low pass filter.

### 3.2.2.2 Compare fall detection algorithms

We wanted a fall detection to be calculated solely on the device. Therefore, we used a threshold algorithm that is not complicated to compute. We compared the accuracy of the following 3 threshold algorithms.

- 2-phase threshold-based of [1]. We tried all possible parameters with the fall dataset to find the best threshold setting (Lower Fall Threshold (LFT) and Upper Fall Threshold (UFT)).

- 3-phase threshold-based of [3]. We tried all possible parameters with the fall dataset to find the best threshold setting (phase 1, phase 2, and the lower and upper bound of phase 3).

- Dynamic Time Warping (DTW). DTW [11] is one way to measure the similarities between two temporal sequences. To use DTW with fall detection, we must find the representative graph for the fall activity and the detection threshold. For each subject, we brought each fall data to calculate the distance with the remaining n-1 subjects using Dynamic Time Warping, summed them together, and then selected the fall data with the smallest total distance. Then, we got the fall activity representative graph of each subject. After that, we chose the global fall activity representative graph by compared each fall activity representative graph of each subject with all data using DTW. We chose the one with the smallest distance. To determine the falling threshold, we applied the one-class classification from the network anomaly detection method shown in Muniyandi *et al.* [9]. They used *k*-Means to partition the training instance into non-overlapping clusters and then used the decision tree to investigate anomaly instances in each cluster. We brought the DTW value of the global fall activity representative graph and all fall data into *k*-Means with $k = 3$. Since the data was distributed in 1 dimension, we used the seam of cluster 2, and 3 as the thresholds instead of using decision trees.

### 3.3 A fall alert system with prior-fall activity identification

At this point, we had a precise model for activity recognition and an accurate fall detection algorithm. We used them to create a fall alert system with prior-fall activity identification. When a fall is detected, the device sends the stored activity data to the server. The server then classifies the activity and notifies the caretaker. The fall alert system overview is shown in Figure 5.

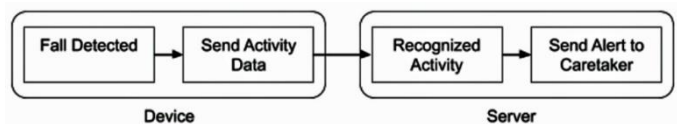

**Figure 5** Overview of our fall alert system with prior-fall activity identification.

We modified a device to record the latest movement data for 4 seconds all the time, which is 200-dimensional data. When a fall is detected, the device will send this 200-dimensional data to the server. Dernbach *et al.* [4] experimented with the size of the window that gave the most accurate activity detection. They found that the window size in 1-2 seconds range gives the highest accuracy. Thus, we set the server to divide data into 100 dimensions into 5 groups, i.e., Data at index $1 - 100, 10 - 110, 20 - 120, 30 - 130$ and $40 - 140$ which are $4 - 2, 3.8 - 1.8, 3.6 - 1.6, 3.4 - 1.4$, and $3.2 - 1.2$ seconds before falling respectively. We used data at index $1–140$ because the data at index 141 – 200 is mostly data of a fall, so we did not make a detection. These 5 groups of data were sent to the classifier to predict activity. After that, the majority vote was used in the response.

## 4. Results

### 4.1 Performance comparison of activity detection models

We used the activity dataset in the experiment to compare the classifiers' accuracy when attaching devices in different locations. There are 14,535 records from arm location, 15,030 records from chest location, 14,143 records from foot location, 16,753 records from leg location, and 12,944 records from wrist location. There are 73,405 records in total. At each location, we also compared data in the Cartesian coordinate system with data in the Spherical coordinate system. Parameters were obtained from Randomized Search using the 5-Fold Cross-Validation method 100 times from the training set. The ratio between the training and the validation set was 70:30. Table 1 shows the accuracy values of each classifier for each sensor location and coordinate system.

**Table 1** Predicting the accuracy of 5 classifiers

| Classifier | right arm | | left chest | | left foot | | left wrist | |
|---|---|---|---|---|---|---|---|---|
| | *Cartesian* | *Spherical* | *Cartesian* | *Spherical* | *Cartesian* | *Spherical* | *Cartesian* | *Spherical* |
| Decision Tree | 71% | 69% | 79% | 77% | 76% | 74% | 73% | 71% |
| *k*-NN | 67% | 58% | 69% | 65% | 75% | 67% | 67% | 57% |
| Naive Bayes | 42% | 40% | 67% | 63% | 56% | 46% | 48% | 42% |
| SVM | 40% | 39% | 40% | 49% | 45% | 41% | 38% | 42% |
| XGBoost | 79% | 77% | 86% | 85% | 85% | 85% | 80% | 80% |

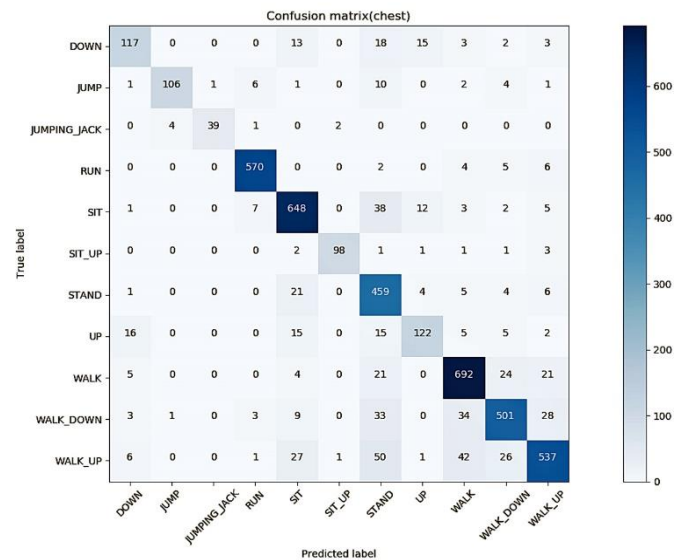

**Figure 6** Confusion matrix of XGBoost in chest location using the Cartesian coordinate system.

The results in Table 1 shows that XGBoost is the most accurate model compared with Decision Tree, $k$-NN, Naive Bayes, and SVM. Also, we found that the activity detection would be most accurate when attaching the device to the left chest and using the Cartesian coordinate system. Moreover, it shows that the accuracy of XGBoost in chest location when the dataset is in the Cartesian coordinate system is slightly more accurate than the Spherical coordinate system. Figure 6 shows the confusion matrix of XGBoost classifier in chest location using the Cartesian coordinate. We chose XGBoost under the Cartesian coordinate system to use as activity recognition in the server from these experimental results. We also found that the locations of the device with high detective accuracy are left chest (pocket), left foot (ankle), left wrist, and right arm (upper arm), respectively. Therefore, we chose the left chest as the location of the device in the fall experiment.

*4.2 Performance comparison of fall detection algorithms*

We evaluated 3 threshold algorithms by accuracy using 64 fall data, 40 fall to their knees first data, and 1,000 random data from non-fall activities. The experimental result of each threshold algorithm is shown in Table 2.

**Table 2** Performance evaluation of 3 threshold algorithms

| Threshold algorithm | Fall type | Fall accuracy | Fall sensitivity | Fall specificity |
|---|---|---|---|---|
| 2-phase | Fall | 75.75% | 95.31% | 74.5% |
| | Fall to their knees first | 72.79% | 97.5% | 71.8% |
| 3-phase | Fall | 88.91% | 73.44% | 89.9% |
| | Fall to their knees first | 91.25% | 77.5% | 91.8% |
| DTW | Fall | 78.57% | 73.44% | 78.9% |
| | Fall to their knees first | 77.79% | 85% | 77.5% |

Since we wanted activities other than fall to be detected as a fall at a low rate in our devices, we selected an algorithm with a highly accurate and specificity (low false-positive rate). Therefore, we chose a 3-phase threshold algorithm of [3] as a fall detection in our devices.

*4.4 Prior-fall activity detection*

The activity recognition and fall detection are combined into our alert system. There were 4 subjects from this research to test the alert system. Each subject performed activities from the same recorded activities, then fall and fall to their knees first 2 times. The results are shown in Table 3. Prior-fall activities that our system can detect fall are WALK, WALK_UP, WALK_DOWN, JUMP, JUMPING_JACK, JUMP, RUN, STAND, and UP. We found that the average accuracy of detecting prior-fall activity is 86.25%.

**Table 3** Prior-fall activity detection results of our proposed method

| Prior-fall Fall Activities | Accuracy of prior-fall activities prediction | False detection results |
|---|---|---|
| WALK | 70% | WALK_UP, WALK_DOWN |
| WALK_UP | 100% | - |
| WALK_DOWN | 100% | - |
| JUMP | 90% | JUMPING_JACK |
| JUMPING_JACK | 100% | - |
| RUN | 100% | - |
| STAND | 70% | SIT |
| UP | 60% | SIT, STAND |
| SIT, SIT_UP, DOWN | - | cannot detect falling |

## 5. Discussion

We found that, in all detection models, most of the incorrect detections are similar. Consider 2 groups of activities; [WALK_UP, WALK_DOWN, WALK] and [STAND, SIT, UP, DOWN]. Most predicting activities are wrong within these groups. We thought that a detection result is in a group of activities because the received activity data, including fall, is collected for 4 seconds, which many activities occur during that period. Therefore, we thought that detecting activities with similar gesture movements is an interesting task.

From the experiment, there are some activities that, when performing, cannot be able to detect falls, that is, SIT, SIT_UP, and DOWN. We thought that the fall could not be detected because the distance between the chest and floor is too close causes the value of G-Force cannot pass the threshold criteria. Most fall detection studies, the subject's body was in a standing position before the fall. Detecting a fall in another pose is also an interesting issue.

## 6. Conclusions

We combined activity detection and fall detection into a fall alert system with prior-fall activity identification in this work. Initially, we made a device to attach to various points on the body and keep records of the activities. We then select the XGBoost prediction model and chest location that brings the most accurate activity detection. We attach the device to that location and test for falls and falls to their knees first. The experimental results show an 86.25% average accuracy when identifying prior-fall activity when a fall occurs. We believe that our alert system could be developed and implemented for elderly monitoring in real environments, allowing the caretakers to manage their responses effectively.

## 7. Acknowledgements

This work was supported by CMU Junior Research Fellowship Program, Chiang Mai University.

## 8. References

[1] Bourke AK, O'Brien JV, Lyons GM. Evaluation of a threshold-based tri-axial accelerometer fall detection algorithm. Gait & Posture. 2007;26(2):194-199.

[2] Cao Y, Yang Y, Liu W. E-FallD: A fall detection system using android-based smartphone. In: 2012 9th International Conference on Fuzzy Systems and Knowledge Discovery; 2012 May 29-31; Sichuan, China. IEEE; 2012. p. 1509-1513.

[3] Chaitep T, Chawachat J. A 3-phase threshold algorithm for smartphone-based fall detection. In: 2017 14th International Conference on Electrical Engineering/Electronics, Computer, Telecommunications and Information Technology (ECTI-CON); 2017 Jun 27-30; Phuket, Thailand. IEEE; 2017. p. 183-189.

[4] Dernbach S, Das B, Krishnan NC, Thomas BL, Cook DJ. Simple and complex activity recognition through smart phones. In: 2012 8th International Conference on Intelligent Environments; 2012 Jun 26-29; Guanajuato, Mexico. IEEE; 2012. p. 214-221.

[5] Ha S, Choi S. Convolutional neural networks for human activity recognition using multiple accelerometer and gyroscope sensors. In: 2016 International Joint Conference on Neural Networks (IJCNN), 2016 Jul 24-29; Vancouver, BC, Canada. IEEE; 2016. p. 381-388.

[6] Lau SL, König I, David K, Parandian B, CariusDüssel C, Schultz M. Supporting patient monitoring using activity recognition with a smartphone. In: 2010 7th International Symposium on Wireless Communication Systems; 2010 Sep 19-22; York, UK. IEEE; 2010. p. 810-814.

[7] Lehsan K, Bootkrajang J. Predicting physical activities from accelerometer readings in spherical coordinate system. In: 2017 18th International Conference on Intelligent Data Engineering and Automated Learning (IDEAL), 2017 Oct 30-Nov 1; Guilin, China, Springer, p. 36-44.

[8] Mao A, Ma X, He Y, Luo J. Highly Portable, Sensor-Based System for Human Fall Monitoring. Sensors. 2017;17:2096.

[9] Muniyandi AP, Rajeswari R, Rajaram R. Network anomaly detection by cascading k-means clustering and c4.5 decision tree algorithm. Procedia Engineering. 2012;30:174-182.

[10] Ravi N, Dandekar N, Mysore P, Littman ML. Activity recognition from accelerometer data. In: 2005 17th conference on Innovative applications of artificial intelligence, 2005 Jul 9-13; Pittsburgh, Pennsylvania, USA, AAAI Press, p. 1541-1546.

[11] Senin. Dynamic time warping algorithm review [Internet]. 2008 [cited 2019 Oct 9]. Available from: http://seninp.github.io/assets/pubs/senin_dtw_litreview 2008.pdf

[12] Shoaib M, Bosch S, Incel OD, Scholten H, Havinga PJ. A survey of online activity recognition using mobile phones. Sensors. 2015;15:2059-2085.

[13] Siirtola P, Röning J. Recognizing human activities user-independently on smartphones based on accelerometer data. International Journal of Interactive Multimedia and Artificial Intelligence. 2012;1:38-45.

[14] Wang J, Chen R, Sun X, She MF, Wu Y. Recognizing human daily activities from accelerometer signal. Procedia Engineering. 2011;15:1780-1786.

[15] WHO. WHO global report on falls prevention in older age [Internet]. 2007 [cited 2018 Oct 21].Available from: http://www.who.int/ageing/publications/Falls_prevention7March.pdf

[16] Zhang W, Zhao X, Li Z. A comprehensive study of smartphone-based indoor activity recognition via Xgboost. IEEE Access. 2019;7:80027-80042.